\documentclass[article]{elsarticle}

\usepackage{times}
\usepackage{latexsym}
\usepackage{geometry}

\usepackage{subfiles}
\usepackage{multirow}
\usepackage{array}
\usepackage{hyphenat}
\usepackage{subcaption}
\usepackage{tabularx}
\usepackage{caption}
\usepackage{graphicx}
\usepackage{adjustbox}   
\usepackage{enumitem}
\usepackage{url}

\usepackage[dvipsnames]{xcolor}
\usepackage{bm}
\usepackage{booktabs}
\usepackage{float}
\usepackage{fancyvrb}

\usepackage{soul}
\usepackage{color}

\usepackage[ruled,vlined]{algorithm2e}
\SetAlFnt{\small}
\usepackage{mathtools}
\usepackage{amsmath}
\usepackage{wrapfig}

\usepackage{tabu}
\usepackage{ctable}

\usepackage{placeins}

\usepackage{graphicx} 

\usepackage{multirow}%
\usepackage{amsmath,amssymb,amsfonts}%
\usepackage{amsthm}%
\usepackage{mathrsfs}%
\usepackage[title]{appendix}%
\usepackage{textcomp}%
\usepackage{manyfoot}%
\usepackage{booktabs}%
\usepackage{listings}%
\usepackage{subfiles}
\usepackage{multirow}
\usepackage{booktabs}
\usepackage{makecell}
\usepackage{wrapfig}
\usepackage[math]{cellspace}
\usepackage{url}
\usepackage{changepage}
\usepackage{makecell}
\usepackage[breaklinks]{hyperref}
\usepackage[english]{babel}
\makeatletter
\newcommand{\thickhline}{%
    \noalign {\ifnum 0=`}\fi \hrule height 1pt
    \futurelet \reserved@a \@xhline
}
\newcolumntype{"}{@{\hskip\tabcolsep\vrule width 1pt\hskip\tabcolsep}}
\makeatother

\usepackage{setspace}
\usepackage{lineno}
\modulolinenumbers[5]

\journal{arXiv}
\usepackage{tabularx}

\begin{document}

\begin{frontmatter}

\title{Large language model-based agents for automated research reproducibility: an exploratory study in Alzheimer's Disease}

\author[1]{Nic Dobbins, PhD\fnref{equal1}\corref{cor1}}
\author[1]{Christelle Xiong, BS\fnref{equal1}}

\author[2]{Kristine Lan, MS}
\author[2]{Meliha Yetisgen, PhD}

\fntext[equal1]{N. Dobbins and C. Xiong contributed equally to this work.}
\cortext[cor1]{Corresponding author: Nic Dobbins, PhD, Biomedical \& Health Informatics, University of Washington, Box 358047 Seattle, WA 98109, USA, {nic.dobbins@jhu.edu}}


\address[1]{Biomedical Informatics \& Data Science, Johns Hopkins University, 3400 N. Charles Street
Baltimore, MD 21218, USA}

\address[2]{Biomedical \& Health Informatics, University of Washington, Box 358047 Seattle, WA 98109, USA}


\begin{abstract}

\noindent\textbf{Objective}: To demonstrate the capabilities of Large Language Models (LLMs) as autonomous agents to reproduce findings of published research studies using the same or similar dataset.

\noindent\textbf{Materials and Methods}: We used the "Quick Access" dataset of the National Alzheimer's Coordinating Center (NACC). We identified highly cited published research manuscripts using NACC data and selected five studies appeared reproducible using this dataset alone. Using GPT-4o, we created a simulated research team of LLM-based autonomous agents tasked with writing and executing code to dynamically reproduce the findings of each study, given only study Abstracts and Methods sections and data dictionary descriptions of the dataset.

\noindent\textbf{Results}: We extracted 35 key findings described in the Abstracts across 5 Alzheimer's studies. On average, LLM agents approximately reproduced 53.2\% of findings per study. Numeric values and range-based findings frequently differed between studies and agents. Also, the LLM agents applied statistical methods or parameters that varied from the original studies, though overall trends and significance were sometimes similar.

\noindent\textbf{Discussion}: In certain cases, LLM-based agents accurately replicated research techniques and findings. But in others, they failed because of implementation flaws or missing methodological information. These discrepancies demonstrate the current constraints of LLMs' capacity to completely automate reproducibility evaluations. However, this preliminary investigation highlights the potential of structured agent-based systems to provide scale assessment of scientific rigor.

\noindent\textbf{Conclusion}: This exploratory work demonstrates both the promise and the limitations of LLMs as autonomous agents for automating reproducibility in biomedical research. We demonstrate the potential for LLMs to validate, critique, and potentially reproduce published research.

\end{abstract}


\end{frontmatter}


\section{Introduction}
\label{sec:introduction}

In recent years, Large Language Models (LLMs) have demonstrated remarkable capabilities in many traditional biomedical natural language processing (NLP) tasks, such as information extraction \cite{xu2024large}, ontology generation \cite{babaei2023llms4ol}, concept normalization \cite{dobbins2024generalizable}, as well as relatively recent, emerging tasks such as diagnostic reasoning \cite{savage2024diagnostic}, radiology reporting \cite{elkassem2023potential}, cohort discovery \cite{dobbins2023leafai}, code generation \cite{zhang2024codeagent} and self-reflection and correction \cite{boiko2023autonomous}. These tasks often require high degrees of reasoning, interpretation, and analysis of complex biomedical data. A subset of these tasks \cite{zhang2024codeagent, boiko2023autonomous} also require a reasonable degree of planning, experimentation, engineering, and dealing with ambiguity in data or instructions. Taken together, these findings indicate that, under certain circumstances, modern LLMs can capably perform a number of tasks typically expected of human scientists.

At the same time, the scientific community has been said to be undergoing a "reproducibility crisis" \cite{baker2016reproducibility}, with an estimated 75\% to 90\% of published studies estimated to not be reproducible \cite{begley2015reproducibility}. While the reasons for this are complex and not necessarily indicative of consistently poor study design or falsification of results, the inability of scientists to reproduce others' findings in many cases stands as a clear challenge to future scientific progress. Moreover, within biomedical informatics (as well as other computationally-driven research domains), the process of reproducing published study results - reading a manuscript, acquiring study data, establishing an environment to conduct experiments, and often writing or reviewing computer code following study methods - is highly time- and resource-intensive, precluding large-scale study replication. 

We hypothesized that LLMs may potentially be applied to automate certain aspects of reproducibility and sought to explore their limitations and capabilities in this context. As LLMs model language - literally outputting sequences of text strings (or audio, images or video) - an additional external component is needed to enable interaction with datasets and code execution: the "wrapping" of an LLM for iterative, turn-based interaction and enablement of external tool usage, such as web search, code interpreters, application programming interfaces (APIs) and so on. Models equipped in such a fashion are called "agents" \cite{gao2024empowering, wu2023autogen}.

In this study, we explore the use of an LLM agent-based system for reproduction of highly cited peer-reviewed manuscripts using the National Alzheimer's Coordinating Center's (NACC) "Quick Access" dataset. NACC serves as a central repository and organizing hub for 35 Alzheimer's Disease Research Centers (ADRCs) across the United States which collect detailed, high-quality data from tens of thousands of patients at risk for or with suffering from Alzheimer's Disease (AD) and other neurological disorders. As NACC data have been widely used for research purposes for decades \cite{beekly2007national} and are freely available to research groups around the world, published studies using NACC data lend themselves in many ways to automated reproducibility.

We aimed to explore whether a group of LLM agents assigned various roles and instructed to develop a plan, explore data, and write and execute code could reproduce published study findings on Alzheimer's Disease. All code used in the study as well as transcripts of LLM agent responses, actions and code are available at \url{https://github.com/ndobb/nacc_paper_llm_reproducer}\footnote{Code will be made available upon article acceptance}.

\section{Background}
\label{sec:background}

\subsection{LLM agents}

Autonomous agents have long been a research interest in the artificial intelligence and computer science communities \cite{franklin1996agent, maes1993modeling, wooldridge1995intelligent}. Prior to widespread use of deep learning-based models from the 2010s, agentic systems were often designed as rule-based software programs driven by relatively simple heuristic policy functions aimed at determining an optimal sequence of actions toward some larger goal. Such systems typically depended on hand-written rule curation and syntactic parsing of natural language \cite{katz1999integrating}, limiting their potential utility relative to humans in open-domain settings with evolving, complex objectives and various means of interacting with a given environment. 

More recently, LLMs have been increasingly studied as core engines for goal interpretation, user interaction, planning, action selection, data analysis and other components of potential agentic systems \cite{wang2024survey}. Exploration in the use of LLM agents has been done in various domains, including social simulation of daily life and cognitive development \cite{liu2023chain, park2023generative, kovavc2023socialai}, jurisprudence and legal interpretation \cite{cui2023chatlaw, hamilton2023blind} and research assistants \cite{ziems2024can, bail2024can}. Common features of these LLM agentic systems include capabilities for planning smaller sequences of steps to solve larger, more complex or abstract goals, storage and retrieval of past actions and observations as "memories" to be used in future decision-making, tool intergration (e.g., APIs or knowledge bases) and flexible response capabilities while interacting with a given environment. See Wang \textit{et al} \cite{wang2024survey} for a detailed examination of these trends.

\subsection{Automated Study Reproducibility}

Study replication is typically a highly time- and resource-intensive task, typically done manually. Individual research studies are, by design, generally unique and heterogeneous, necessitating experimental setup parameters, datasets, computing needs and more specific to a given research manuscript or project. While short of full automation, many researchers have nonetheless explored methods, representations, and tools aimed toward facilitating reproducibility. These efforts have tended to focus either on \textit{abstraction}, the expression of methodology and steps in generalizable patterns, and \textit{source code execution} to recreate experiment results in environments outside of the originating lab. In the biochemical and materials science domain, Canty and Abolhasani \cite{canty2024reproducibility} explored abstraction methods and representations from computer science for representing chemical reactions and processes. Beaulieu-Jones and Greene \cite{beaulieu2017reproducibility} developing "Continuous Analysis", a reproducibility framework for computation-based experiments leveraging containerization and virtualization technologies. Harwell and Gini \cite{harwell2022sierra} similarly developed SIERRA, a programmatic framework for experimental parameter management and injection for study reproduction. 

Other studies have evaluated the use of LLM agents for independent hypothesis generation, experiment planning, and execution. Boiko \textit{et al} \cite{boiko2023autonomous} created an agentic system around OpenAI's GPT-4 as a "Coscientist" capable of internet and document search, code generation and execution. The authors also note Coscientists's capabilities for hypothesis generation and experiment automation, though not specifically for reproducibility. Most closely related to and as a strong complement to our study, Ifargan \textit{et al} \cite{ifargan2025autonomous} demonstrate a proof-of-concept LLM agent-based framework to autonomously generate research manuscripts from LLM-generated hypotheses in an end-to-end fashion. The authors note that the agent-generated manuscripts were generally without major errors in 80-90\% of cases. Additionally, by tracking information flow, their platform creates “data-chained” manuscripts, in which downstream results are linked to upstream code and data, thus setting a new standard for the verifiability of scientific outputs.
The progress made in automation for scientific research is significant, but the challenge of automating the reproducibility of prior studies, which involves faithfully re-executing and validating them, remains relatively uncharted territory. This study aims to build upon previous work by exploring whether large language model-based autonomous agents can dynamically reproduce findings from real-world, peer-reviewed research studies, utilizing only publicly available datasets and minimally curated prompts.

\subsection{Key Contributions}

To the best of our knowledge, no studies to-date have evaluated the use of LLM agentic systems for automating published study reproducibility. Our study makes the following key contributions:

\begin{enumerate}
    \item A methodology and experimental setup and evaluation for use of LLM agents to reproduce published journal articles.
    \item A case study in Alzheimer's Disease using 5 highly-cited peer-reviewed journal articles with 25/35 (71.4\%) of key findings automatically reproduced.
    \item A detailed error analysis and evaluation of statistical and other methodological divergences for guiding future studies. 
\end{enumerate}

\section{Methods}
\label{sec:methods}

\subsection{Dataset}

Data for this study came from the NACC "Quick Access" Uniform Dataset (UDS) version 3, downloaded in spring of 2023. NACC data are updated and released at regular intervals as data are collected and harmonized among 35 ADRCs \cite{besser2018version}. Study approval was obtained from Institutional Review Boards at each respective ADRC. 

NACC data consist of multiple "modules", the base dataset of which is the UDS, composed of longitudinal clinical data for over 40,000 participants. Additional modules also exist, including FTLD (frontotemporal lobar degeneration), LBD (Lewy body disease), neuropathology (autopsy), imaging, and more. For this initial study, we intentionally selected only the UDS "Quick Access" dataset, believing that incorporating additional files and datasets would unnecessarily complicate the study design. We felt that the base UDS dataset alone was more suitable for developing our methodological approach and establishing an evaluation baseline.

\subsection{Studies}

\begin{figure}[H]
      \centering
      \includegraphics[scale=0.5]{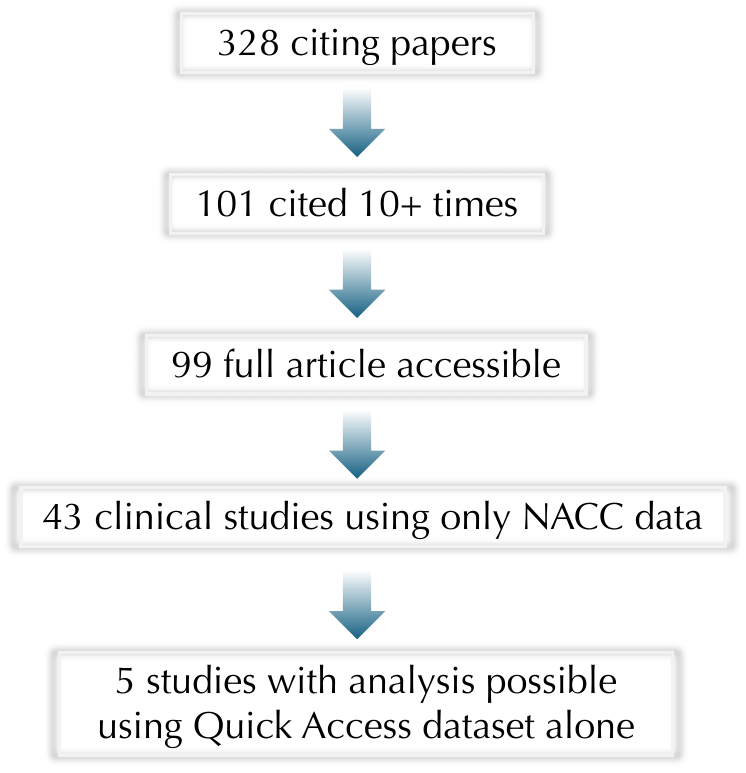}  
    \caption{Process for determining eligible studies for inclusion. Our initial review of citing papers for the NACC version 3 manuscript was conducted in Google Scholar on November 6, 2024.}
    \label{fig_studies}
\end{figure}

Our process for determining eligible studies is outlined in Figure \ref{fig_studies}. One challenge in our study design arose from the fact that the NACC dataset is continuously updated and we did not have access to historical versions. As a result, we assumed that our local dataset would be structurally identical (in schema) to more recent studies, but would likely contain slightly more data rows, as it included records for participants seen since the publication of earlier studies. While some differences in the data were unavoidable, we aimed to minimize these discrepancies by initially filtering for studies that cited the 2018 publication on the version 3 update to the NACC UDS schema \cite{besser2018version}, reasoning that these studies would likely be using a version of the dataset similar to ours.

\subsection{Study Reproduction Process}

Our process for determining research studies we could replicate was as follows:

\begin{enumerate}
    \item We conducted a Google Scholar search \footnote{\url{https://scholar.google.com/scholar?cites=15070434224497513322&as\_sdt=5,48&sciodt=0,48&hl=en}} on November 6, 2024 for research studies citing that manuscript, which returned 328 total.
    \item We reasoned that highly-cited studies would be more important to aim to replicate, as they in turn likely influenced further studies. We thus filtered to only those cited 10 times or more, which yielded 101.
    \item 99 articles had full texts full accessible.
    \item We excluded studies which were not clinical research papers, did not use NACC data, or used NACC data but with other datasets as well, yielding 43.
    \item We then excluded studies which we determined likely did not use the UDS Quick Access dataset alone, which yielded 5 studies \cite{kiselica2020development, kiselica2020using, kiselica2023examining, saari2020network, kurasz2020ethnoracial}.
\end{enumerate}

After selecting the five studies to attempt to reproduce, we extracted the text of their abstracts and methods sections into separate text files. Figures and tables were excluded from these files, and we explicitly assumed that the information contained in them was not essential for reproducing the methods, though we recognize that this may not always be the case. To the best of our knowledge, none of these studies had made their code publicly available (e.g., via a GitHub repository).

Finally, we conducted a detailed review of each study and determined the most likely NACC Quick Access columns used or potentially useful for analysis. These columns of interest were in used as part of our later prompting process.

We used the Autogen \cite{wu2023autogen} version 2 library in Python and OpenAI's GPT-4o model \cite{hurst2024gpt} to conduct our experiments. Autogen is an open-source framework for managing autonomous LLM-based agents. We chose GPT-4o as it was among the best performing widely-available "flagship" LLM models at the time of our experiments. 

\begin{figure}[H]
      \centering
      \includegraphics[scale=0.53]{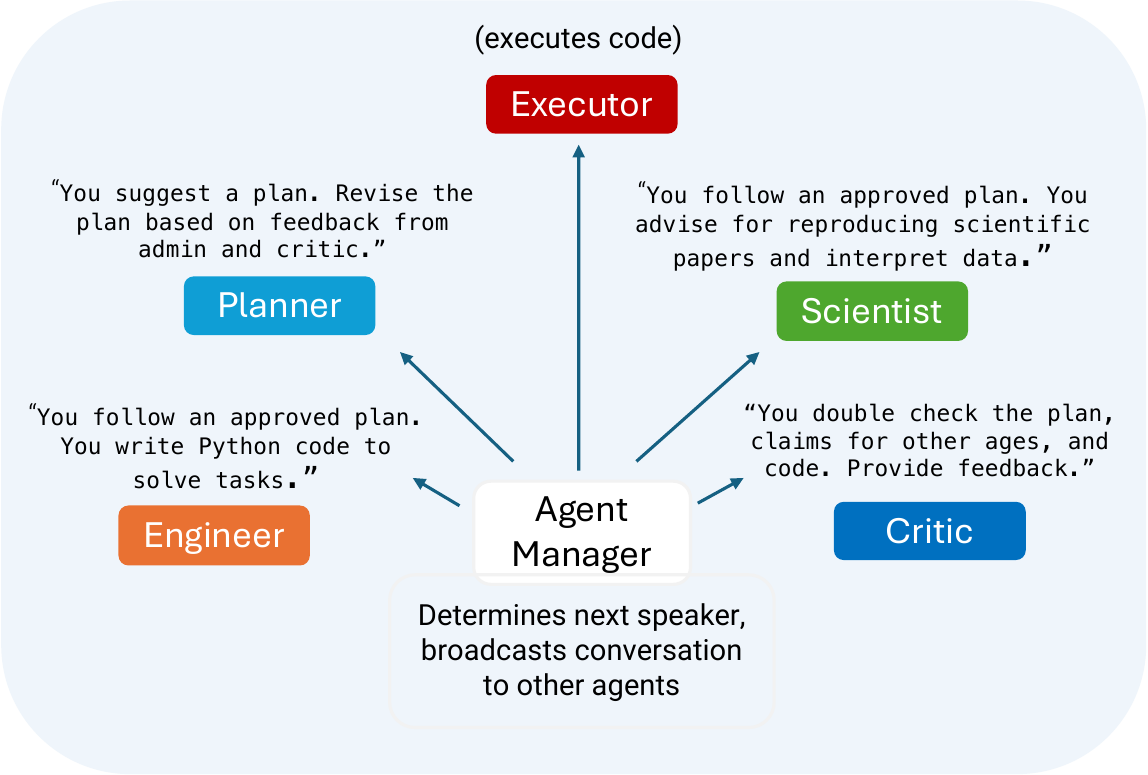}  
    \caption{Agent roles in this study. Individual agent instructions (shown in quotes) are abbreviated for readability. The entire agent team is given a single set of instructions by the user, after which the agent manager orchestrates their iterative response.}
    \label{fig_agents}
\end{figure}

The Autogen framework can be used by declaring agent team roles - literally free-text descriptions of the background, perspectives, and goals of each agent - as well as any tool usage and code execution capabilities they should be allowed. Finally, an agent "manager" is declared, which facilitates agent-to-agent communication and chooses the next appropriate agent as "speaker" to conduct an action or respond to a stimuli. As this was an exploratory study, we based our agent role definitions largely on example team compositions provided by the Autogen team \footnote{\url{https://github.com/microsoft/autogen/tree/0.2/notebook}} as a starting point. We declared five agent roles, shown in Figure \ref{fig_agents}:

\begin{enumerate}
    \item \textbf{Planner} - suggests and revises a plan for the agent team to follow based on instructions from the user.
    \item \textbf{Engineer} - follows the plan and writes Python code.
    \item \textbf{Scientist} - follows the plan, advises on strategies for reproducing research paper results and interpreting code output.
    \item \textbf{Critic} - critiques the plan and provides feedback for improvement.
    \item \textbf{Executor} - a special non-LLM agent which executes Python code passed from the Engineer.
    \item \textbf{Manager} - a special agent which orchestrates the team, chooses speakers, and broadcasts speaker responses to other agents to track cumulative actions and responses.
\end{enumerate}

We developed our initial prompting strategy using a single research study \cite{kapoor2024robust}, which used NACC Quick Access UDS data. We identified this study through a keywork search for recent NACC-based AD research and selected it intentionally as a development set for prompt design. We chose this study because we believed it applied relatively straightforward statistical methods, such as linear regression analysis, which we believed would make it easier to replicate with LLM agents. Using this study, we iteratively refined the structure, scope, and role assignments of our agent prompts before applying the final prompting strategy to the five main studies in our analysis.

\begin{figure}[H]
      \centering
      \includegraphics[scale=0.53]{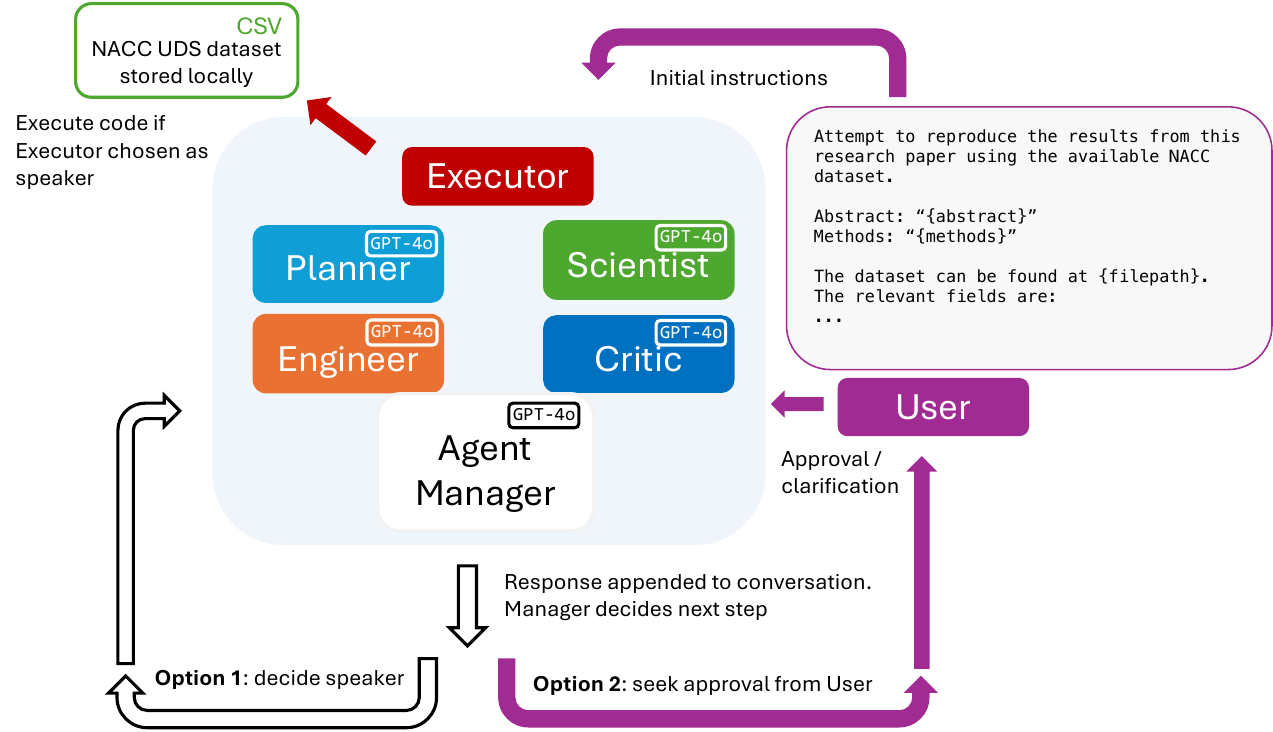}  
    \caption{Process for prompting agents and progressing the conversation simulation.}
    \label{fig_process}
\end{figure}

The process for prompting our LLM agent team and iteratively reproducing study results is depicted in Figure \ref{fig_process} and is as follows:

\begin{enumerate}
    \item We manually generate an initial instruction prompt for each project, which explained the task of reproducing a given study's results using the NACC dataset. This prompt included the full Abstract and Methods sections from the original manuscript. Also, the initial prompt included manually selected variables from the NACC data dictionary relevant for replicating the study (typically 15–30 columns per study, depending on complexity). In addition to these variable definitions, we included dataset-level instructions, such as how to interpret date formats, how to handle missing values, and specific reminders to apply cohort inclusion and exclusion criteria as described in the original study before performing any analysis. This prompt served as the input for the “Admin Agent” to initiate the replication workflow. The full reproduction prompt, including study details, data dictionary, and analysis instructions, is provided in Appendix~\ref{app:prompt}.

    We specifically instructed the agents to \textbf{recreate only results reported in the abstract}. We selected this approach because abstracts typically highlight the study’s most important and relevant findings, and attempting to replicate every result from the full text would be infeasible at scale. To define the target outputs for reproduction, we manually extracted and listed the key statistical assertions described in each study’s abstract. These manually curated assertions served as the basis for evaluating the agents’ performance.
    \item For each subsequent step, the agent manager chooses either an agent to respond or perform an action or seeks clarification or approval from the user. We enabled the agents to seek user approval at key decision points to keep the user in the loop and ensure transparency. This design choice was informed by patterns observed in Autogen examples, where user-in-the-loop approval is frequently emphasized as a best practice.
    \item If user approval was sought, we responded by reinforcing the need to use appropriate columns and strictly follow study methods, then re-asserted the abstract contents. Alternatively if the agents sought approval to complete the exercise but parts of abstract results had not yet be attempted, we responded by instructing the LLMs to specifically attempt to recreate those remaining findings.
\end{enumerate}

\subsection{Evaluation}

Evaluating whether a simulated research team of LLM agents is able to successfully recreate results reported in an abstract is a challenging and somewhat open-ended problem. We determined that a reasonable starting point for such an analysis would be to extract individual assertions from results sections of each abstract as separate data points, then evaluate accuracy as a primary metric of interest.

For example, \cite{kurasz2020ethnoracial} includes in the abstract results the sentences "Presentation age was similar between groups (median 74–75 years). Compared to Whites (n=1782), African Americans (n=130) and Hispanics (n=122) were more likely to be female and single." From this we extracted the following individual assertions:

\begin{enumerate}
    \item Median presentation age (White) - 74-75
    \item Median presentation age (African American) - 74-75
    \item Median presentation age (Hispanic) - 74-75
    \item Black more likely female - True
    \item Hispanic more likely female - True
    \item Black more likely single - True
    \item Hispanic more likely single - True
\end{enumerate}

As we knew our dataset to be slightly newer than the NACC dataset used in studies, we did not include study-reported denominators of patients as assertions to match and assumed in nearly all cases precise numbers of patients would be slightly different.

For each of the five studies, we manually reviewed the abstract to identify key findings, then executed our study reproduction steps accordingly. We subsequently evaluated whether the agent’s outputs aligned with each of these findings. A summary of alignment for each key finding across studies is presented in Table~\ref{tbl_methods}, with a checkmark indicating successful reproduction. We considered a finding aligned if, if a mean, median, or percentage, agent findings were within a value of 1. For other assertions, typically Boolean, we considered findings aligned if both were true or both were false, though we acknowledge such a match may not indicate true study replication in certain cases (if underlying values behind the assertion differed).

\section{Results}
\label{sec:results}

Results are shown in Table \ref{tbl_results}. In 3/5 (60\%) studies, a majority of findings were approximately replicated by the LLM agents. In 0/5 studies, however, were statistical methods used by LLM agents exactly the same as those described in the study. In two studies \cite{kiselica2020development, kiselica2023examining}, specifics of statistical methods used were largely omitted, instead references to previous studies which detailed were given. In these cases the LLM agents did not thus have access to detailed methods descriptions within our prompt, though it is possible the cited past studies were used as training data for the GPT-4o model. 

\begin{table}[h!]
    \small
    \centering
    \caption{Comparisons of 5 studies results presented in original articles and produced by LLM-based agents. Agent outputs considered aligned with study findings are highlighted in bold. Numeric results often differed from study findings, while Boolean assertions more frequently aligned. Statistical methods employed by LLM agents in each study often differed in subtle and unsubtle ways from those described in study Methods.}

\begin{adjustwidth}{0cm}{}
\scalebox{0.95}{
\begin{tabular}{l|llll}
     & & \multicolumn{2}{c}{\makecell{\textbf{Findings}}} & \\
    \textbf{Study Name} & \textbf{Assertion} & \textbf{Study} & \textbf{Agents} \\
    \toprule
    \multirow{3}{*}{
        \makecell{
            Using multivariate... \cite{kiselica2020using}
        }
    }
        & MBR of low scores range & 1.40 to 79.2\% & 10.9 to 51.3\%  \\
        & Posttest probabilities at 2 year& 0.06 to 0.33 & 0.04 & \\
        & Posttest probabilities conversion to MCI & 0.12 to 0.32 & 0.67 & \\
    \hline
    \multirow{5}{*}{
        \makecell{
            Network structures... \cite{saari2020network}
        }
    } \\
        & Mean follow up time in days & 387 & 411 \\
        & Informant symptoms lack of pos. affect & True & \textbf{True} & \\
        & Network structure p-value	& 0.71 & 0.42 & \\
        & Network connectivity p-value & 0.92 & 1 & \\
        & Symptoms clusters formed & 4 & 14 & \\
    \hline
    \multirow{15}{*}{
        \makecell{
            Ethnoracial differences... \cite{kurasz2020ethnoracial}
        }
    } \\
        & Median presentation age &	74 to 75 & \textbf{73 to 74.5} \\
        & Black and Hispanic more likely female & True & \textbf{True} \\
        & Black and Hispanic more likely single & True & \textbf{True} \\
        & Black and Hispanic more likely less education & True & \textbf{True} \\
        & Black and Hispanic more likely cardiov. risk & True & \textbf{True} \\
        & Black and Hispanic likely less medication use & True & \textbf{True} \\
        & Black and Hispanic lower on cogn. tests & True & \textbf{True} \\
        & Hispanic more depressive symptoms & True & \textbf{True} \\
    \hline
    \multirow{7}{*}{
        \makecell{
            Examining racial... \cite{kiselica2023examining}
        }
    } \\
        & Whites higher rate low scores: memory & True & True \\
        & Whites higher rate low scores: attention & True & \textbf{True} \\
        & Whites higher rate low scores: processing speed & True & \textbf{True} \\
        & Whites higher rate low scores: verbal fluency & True & \textbf{True} \\
        & Whites meeting actuarial criteria for MCI & 71.6 & 48.9 \\
        & Blacks meeting actuarial criteria for MCI & 57.9 & 41.3 \\
        & Whites more likely meet actuarial criteria & True & \textbf{True} \\
    \hline
    \multirow{4}{*}{
        \makecell{
            Development and... \cite{kiselica2020development}
        }
    } \\
        & SRB equations predict cognitive variables & True & \textbf{True} \\
        & Base rate of at least 1 SRB decline range & 26.7 to 58.1 & 5.65 to 52.0 \\
        & SRB indices for measures showed assoc. & True & \textbf{True} \\
        & CDR-B impairment associated with decline & True & \textbf{True} \\
\end{tabular}
}\end{adjustwidth}
    
    \label{tbl_results}
\end{table}

Table \ref{tbl_methods} summarizes the alignment comparisons between the original study methods and the implementations by the LLM-based agents in these five studies. In the 8/14 method comparisons,  LLM agents reproduced the original statistical methods justifiably. These 8 methods are general statistical analysis, such as chi-square tests, t-test, and z-scores, while the remaining 6 methods are using specific statistical methods relevant to specific areas, such as demographic-corrected regression analysis in psychology and neuroscience. 

While the general statistical methods were often implemented correctly, the domain-specific methods faced challenges for LLM-based agents. To better understand how these mismatches occurred, we qualitatively analyzed the agent behavior in response to these domain-specific methods without giving additional information in prompts. LLM-based agents handled the six methods using two strategies. In some cases, such as, network robustness and community detection, the agents just skipped and ignored the implementation of this task. In other cases, the agent attempted to approximate the method using general statistical methods rather than domain-specific methods. For example, demographic-corrected regression analysis is a specific method in psychology, which adjusts for key covariates, such as age, sex, and education. We did not provide the specific-domain knowledge in prompt. In addition, no specific context was mentioned in the original study method section. During reproduction, LLM-based agent only implemented a basic, unadjusted regression model, omitting the demographic adjustments.

\begin{table}[h!]
    \small
    \centering
    \caption{Comparisons of study methods and LLM-based agent implementations, with justification of findings based on method alignment.}
    \begin{scriptsize} 
\begin{center}

\begin{tabularx}{\textwidth}{l|X X X}
    \toprule
    \textbf{Study Name} & \textbf{Study Methods} & \textbf{Agent Methods} & \textbf{Findings Justifiable by Agent Methods} \\
    \midrule
    
    \multirow{2}{*}{
        \makecell[l]{Using multivariate... \cite{kiselica2020using}}
    } 
        & Demographic-corrected regression analysis & Unadjusted regression analysis & Not justifiable. Agent omitted key covariates (sex, age, and education), invalidating the regression. \\
        & Posttest probability across different subgroups & Posttest probability for the full study population & Not justifiable. Agent failed to stratify by subgroups, returning a single value instead of a range.\\
        
    \midrule
    \multirow{5}{*}{
        \makecell[l]{Network structures... \cite{saari2020network}}
    } 
        & Network estimation & Same & \textbf{Justifiable}. Agent method matched the original approach. \\
        & Network inference & Same & \textbf{Justifiable}. Agent replicated the original statistical logic. \\
        & Network robustness (bootstrapping) & Not implemented & Not justifiable. Robustness assessment was missing, limiting the validity of conclusions.\\
        & Network comparison & Permutation method & \textbf{Justifiable}. Permutation was an acceptable substitute for the original R package (NetworkComparisonTest). \\
        & Community detection & Not implemented & Not justifiable. Method was essential to original analysis but absent in the agent implementation.\\
        
    \midrule
    \multirow{2}{*}{
        \makecell[l]{Ethnoracial differences... \cite{kurasz2020ethnoracial}}
    } 
        & Kruskal-Wallis and Pearson chi-square tests & Same & \textbf{Justifiable}. Agent accurately reproduced original statistical methods. \\
        & Linear regression (reference group = White) & Linear regression (reference group = Black) & Not justifiable. Change in reference group led to numerically inconsistent results.\\
        
    \midrule
    \multirow{3}{*}{
        \makecell[l]{Examining racial... \cite{kiselica2023examining}}
    } 
        & Chi-square and t-tests & Same & \textbf{Justifiable}. Agent correctly followed original method. \\
        & Demographically adjusted z-scores & Unadjusted z-scores & Not justifiable. Omission of demographic adjustment undermined validity. \\
        & Evaluation of 12 cognitive tests & Evaluation of 6 cognitive tests & Not justifiable. Reduction in test set compromised methodological fidelity. \\
        
    \midrule
    \multirow{3}{*}{
        \makecell[l]{Development and... \cite{kiselica2020development}}
    }
        & Regression-based z-score & Same & \textbf{Justifiable}. Agent replicated the original method. \\
        & Z-score cut-points (two-tailed) & Same & \textbf{Justifiable}. Agent correctly implemented the cut-point strategy. \\
        & Convergent validity testing & Same & \textbf{Justifiable}. Agent reproduced the validation approach. \\
    
    \bottomrule
\end{tabularx}
\end{center}
\end{scriptsize}
    
    \label{tbl_methods}
\end{table}

\begin{figure}[h]
      \centering
      \includegraphics[scale=0.6]{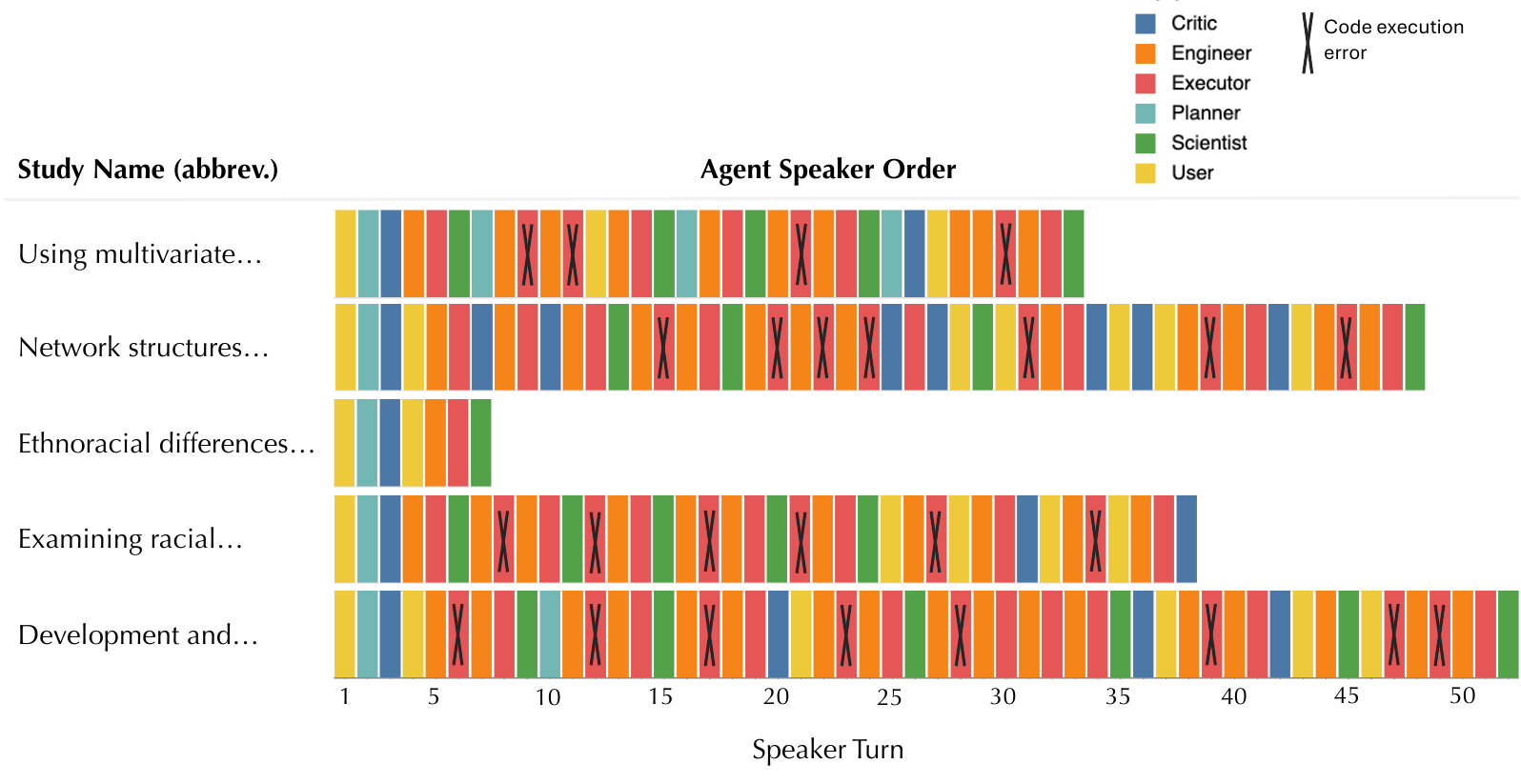}  
    \caption{Sequence of speakers for each study replication attempt. }
    \label{fig_speaker_order}
\end{figure}

Figure \ref{fig_speaker_order} shows the sequence of agent speakers for each study replication attempt. The number of LLM speaking iterations ranged from 7 to 52 (mean: 35.6, stdev: 17.7). Python code executions failed in 26/55 (47.2\%) of attempts, often due to incorrect column names in (e.g. "AGE" rather than "NACCAGE"), despite column names having been provided in the original prompt. In all cases however subsequent corrected Python code executions succeeded.

\section{Discussion}
\label{sec:discussion}

This study demonstrates both the viability and challenges of using LLM-based autonomous agents for reproducing peer-reviewed published study results. Fully replicating study findings proved challenging, and in each replication attempt the LLM agents' statistical methods used varied somewhat from those in studies. In certain cases, such as posttest probability across different subgroups versus for the entire population \cite{kiselica2020using}, these differences were great enough that it would be unreasonable to draw the same conclusions as those of the original study. In other cases, such as permutation methods \cite{saari2020network}, statistical method differences were less impactful and more benign,  ultimately aligning with study results, despite differences in approach.

In a number of cases, for example, demographic-adjusted versus unadjusted regression analysis \cite{kiselica2020using, kiselica2023examining}, differences in statistical methods may have stemmed from the original studies referencing prior publications for methodological details, rather than fully restating them. Our study did not attempt to include details of referenced works in the prompt, and thus it highly possible that the LLM agents were less likely to replicate study results due to lack of detailed information, as opposed to lack of capabilities of the underlying LLM model (GPT-4o). 

Despite these challenges, this work also highlights the capabilities of LLMs to potentially reproduce results of published studies. While statistical results often differed, in many cases the LLM agents were able to iteratively successfully dynamically analyze a new dataset, react to stimuli and code outputs, correct error code, and reason as to possible challenges or approaches and adjust their methods independent of human input. Given the relative simplicity of our approach - using a predefined set of agents and roles which may not be optimal to this task, using a single instruction prompt with not capability for the agents to explore other reference studies, and so on - it is very possible that more sophisticated future studies may build upon our work and demonstrate improved performance.

Importantly, this study strongly underscores the need for nuanced interpretation and close examination of statistical methods necessary when using LLMs for data analysis. As the number of commercial and scientific applications which enable LLMs to directly interact with data, conduct analysis, and report results following human instructions grows, in many cases so to does the peril of trusting LLM-implemented analyses and interpretations without verifying their correctness. 

\subsection{Limitations}

This work was not without limitations. First, we evaluated only five studies for replications, all within the domain of Alzheimer's Disease. Expanded evaluations of a larger sample size of published studies and across different disciplines are necessary to generalize our findings. To the best of our knowledge, the computer code used for these studies is not publicly available, though it is possible that some components are available elsewhere on the Internet and were included in GPT-4o's training data. While the LLM agents used statistical techniques that differed from those in the original publications, these differences are unlikely to have led to unjustified improvements in the results.

This work was further limited to use of only a single language model (GPT-4o) and used a predefined set of agent roles which may not be optimal configuration for this task. The performance of other models, including fine-tuned ones for biomedical applications, as well as alternative roles and approaches for agent assignments, remains uncertain. Nevertheless, given the exploratory nature of this work, we believe our chosen approach and assumptions provide a reasonable foundation for future investigation.

Importantly, we also evaluated only those findings described in study Abstracts, and not the exhaustive findings described in greater detail in study Results sections. Given the exploratory nature of this work we believe this to be a reasonable choice for evaluation, though acknowledge that such an approach is insufficient to enable any conclusions as to LLM agent capabilities to fully replicate published studies end-to-end. Moreover, as many studies describe Abstract results in Boolean expressions, rather than numerically, statistical differences in findings may be appear aligned simply because a similar trend is observed, even if values found by studies versus the LLM agents differ. However, given that our dataset is somewhat newer than that used in the studies, it is also possible that certain differences may arise simply due to differences in the dataset, and may not easily be controlled for. Additionally, to our knowledge there exists no common benchmarks or consensus approach in evaluation for the use of LLMs in reproducing study results, and as such believe the exploratory methods described here to be a valuable, reasonable starting point which future research can expand upon and refine.

Another significant limitation is related to the design of the prompts. In this work, LLM agents were only fed in with the Abstract and Methods sections of each single publication. Lack of access to the original tables and figures included in the Results section. This may affect LLM agents capability of reproducing higher accuracy and highly aligned methods and results, particularly in methods with specific area knowledge and subgroups analysis. Further research may benefit from incorporating structured prompts including both textual descriptions and key tables and figures templates.

Another set of constraints relates to the breadth and uniformity of the reproduced findings. Some of the LLM-generated outputs aligned with the numerical values mentioned in the Methods section, while others only matched the directional discoveries from the Results. This inconsistency reflects a more widespread issue: the models often failed not because of incorrect code, but due to a lack of contextual memory, overlooking previously defined variables or inputs when generating subsequent analyses. These breakdowns in continuity underscore the difficulties in applying stateless, step-by-step prompting to tasks that necessitate coherent, multi-step reasoning and execution.

The agent team roles had practical constraints. For instance, the "Scientist" agent role mainly interpreted results but lacked the ability to evaluate whether the statistical methods used were appropriate. Furthermore, our binary evaluation system, categorizing agent-generated methods as either "same" or "not justifiable", did not account for legitimate but alternative analytic approaches. In one case, a permutation test was substituted for an unavailable R package, and though the form was different, it still met the study's analytic intent. Future frameworks should better differentiate between deviations that are methodologically invalid versus those that are valid but distinct.

Finally, numerical comparisons between agent and study results were sometimes complicated by differences in the study cohort. Agents did not construct cohorts using explicit criteria from the original paper but instead made assumptions based on data availability and variable names. In the results, final values may have involved different populations, lowering the current capability of producibility. A promising future direction may involve allowing agents to explore the dataset freely, identify plausible cohort definitions that approximate the reported study sample, and transparently report the differences. This could enable more qualitative, human-interpretable comparisons and expose discrepancies that are otherwise masked by numeric matching alone.

\section{Conclusion}
\label{sec:conclusion}

The study describes an exploratory approach in the use of LLM-based autonomous agents for dynamically, iteratively recreating the results of published studies on Alzheimer's Disease. While using somewhat different statistical techniques, in 3 of 5 studies the LLM agents were able to largely approximate study findings and trends. We hope this exploratory work serves as a baseline and starting point for future studies in the use of LLMs to replicate scientific studies.

\section{Author contributions}
ND drafted the initial manuscript and executed experiment code. CX and KL performed analysis of the agents' statistical methods. CX led revision of the manuscript and error analysis. MY advised on study design and planned experiment methods. All authors contributed to the manuscript and approved the final version of the work. 

\section{Funding}
The NACC database is funded by NIA/NIH Grant U24 AG072122. NACC data are contributed by the NIA-funded ADRCs: P30 AG062429 (PI James Brewer, MD, PhD), P30 AG066468 (PI Oscar Lopez, MD), P30 AG062421 (PI Bradley Hyman, MD, PhD), P30 AG066509 (PI Thomas Grabowski, MD), P30 AG066514 (PI Mary Sano, PhD), P30 AG066530 (PI Helena Chui, MD), P30 AG066507 (PI Marilyn Albert, PhD), P30 AG066444 (PI David Holtzman, MD), P30 AG066518 (PI Lisa Silbert, MD, MCR), P30 AG066512 (PI Thomas Wisniewski, MD), P30 AG066462 (PI Scott Small, MD), P30 AG072979 (PI David Wolk, MD), P30 AG072972 (PI Charles DeCarli, MD), P30 AG072976 (PI Andrew Saykin, PsyD), P30 AG072975 (PI Julie A. Schneider, MD, MS), P30 AG072978 (PI Ann McKee, MD), P30 AG072977 (PI Robert Vassar, PhD), P30 AG066519 (PI Frank LaFerla, PhD), P30 AG062677 (PI Ronald Petersen, MD, PhD), P30 AG079280 (PI Jessica Langbaum, PhD), P30 AG062422 (PI Gil Rabinovici, MD), P30 AG066511 (PI Allan Levey, MD, PhD), P30 AG072946 (PI Linda Van Eldik, PhD), P30 AG062715 (PI Sanjay Asthana, MD, FRCP), P30 AG072973 (PI Russell Swerdlow, MD), P30 AG066506 (PI Glenn Smith, PhD, ABPP), P30 AG066508 (PI Stephen Strittmatter, MD, PhD), P30 AG066515 (PI Victor Henderson, MD, MS), P30 AG072947 (PI Suzanne Craft, PhD), P30 AG072931 (PI Henry Paulson, MD, PhD), P30 AG066546 (PI Sudha Seshadri, MD), P30 AG086401 (PI Erik Roberson, MD, PhD), P30 AG086404 (PI Gary Rosenberg, MD), P20 AG068082 (PI Angela Jefferson, PhD), P30 AG072958 (PI Heather Whitson, MD), P30 AG072959 (PI James Leverenz, MD).

\section{Conflicts of interest}
The authors declare that there are no competing interests.

\section{Data availability}
There are no new data associated with this article.

\newpage
\bibliography{main}

\begin{thebibliography}{10}

\bibitem{xu2024large}
Derong Xu, Wei Chen, Wenjun Peng, Chao Zhang, Tong Xu, Xiangyu Zhao, Xian Wu, Yefeng Zheng, Yang Wang, and Enhong Chen.
\newblock Large language models for generative information extraction: A survey.
\newblock {\em Frontiers of Computer Science}, 18(6):186357, 2024.

\bibitem{babaei2023llms4ol}
Hamed Babaei~Giglou, Jennifer D’Souza, and S{\"o}ren Auer.
\newblock Llms4ol: Large language models for ontology learning.
\newblock In {\em International Semantic Web Conference}, pages 408--427. Springer, 2023.

\bibitem{dobbins2024generalizable}
Nicholas~J Dobbins.
\newblock Generalizable and scalable multistage biomedical concept normalization leveraging large language models.
\newblock {\em arXiv preprint arXiv:2405.15122}, 2024.

\bibitem{savage2024diagnostic}
Thomas Savage, Ashwin Nayak, Robert Gallo, Ekanath Rangan, and Jonathan~H Chen.
\newblock Diagnostic reasoning prompts reveal the potential for large language model interpretability in medicine.
\newblock {\em NPJ Digital Medicine}, 7(1):20, 2024.

\bibitem{elkassem2023potential}
Asser~Abou Elkassem and Andrew~D Smith.
\newblock Potential use cases for chatgpt in radiology reporting.
\newblock {\em American Journal of Roentgenology}, 221(3):373--376, 2023.

\bibitem{dobbins2023leafai}
Nicholas~J Dobbins, Bin Han, Weipeng Zhou, Kristine~F Lan, H~Nina Kim, Robert Harrington, {\"O}zlem Uzuner, and Meliha Yetisgen.
\newblock Leafai: query generator for clinical cohort discovery rivaling a human programmer.
\newblock {\em Journal of the American Medical Informatics Association}, 30(12):1954--1964, 2023.

\bibitem{zhang2024codeagent}
Kechi Zhang, Jia Li, Ge~Li, Xianjie Shi, and Zhi Jin.
\newblock Codeagent: Enhancing code generation with tool-integrated agent systems for real-world repo-level coding challenges.
\newblock {\em arXiv preprint arXiv:2401.07339}, 2024.

\bibitem{boiko2023autonomous}
Daniil~A Boiko, Robert MacKnight, Ben Kline, and Gabe Gomes.
\newblock Autonomous chemical research with large language models.
\newblock {\em Nature}, 624(7992):570--578, 2023.

\bibitem{baker2016reproducibility}
Monya Baker.
\newblock Reproducibility crisis.
\newblock {\em nature}, 533(26):353--66, 2016.

\bibitem{begley2015reproducibility}
C~Glenn Begley and John~PA Ioannidis.
\newblock Reproducibility in science: improving the standard for basic and preclinical research.
\newblock {\em Circulation research}, 116(1):116--126, 2015.

\bibitem{gao2024empowering}
Shanghua Gao, Ada Fang, Yepeng Huang, Valentina Giunchiglia, Ayush Noori, Jonathan~Richard Schwarz, Yasha Ektefaie, Jovana Kondic, and Marinka Zitnik.
\newblock Empowering biomedical discovery with ai agents.
\newblock {\em Cell}, 187(22):6125--6151, 2024.

\bibitem{wu2023autogen}
Qingyun Wu, Gagan Bansal, Jieyu Zhang, Yiran Wu, Shaokun Zhang, Erkang Zhu, Beibin Li, Li~Jiang, Xiaoyun Zhang, and Chi Wang.
\newblock Autogen: Enabling next-gen llm applications via multi-agent conversation framework.
\newblock {\em arXiv preprint arXiv:2308.08155}, 2023.

\bibitem{beekly2007national}
Duane~L Beekly, Erin~M Ramos, William~W Lee, Woodrow~D Deitrich, Mary~E Jacka, Joylee Wu, Janene~L Hubbard, Thomas~D Koepsell, John~C Morris, Walter~A Kukull, et~al.
\newblock The national alzheimer's coordinating center (nacc) database: the uniform data set.
\newblock {\em Alzheimer Disease \& Associated Disorders}, 21(3):249--258, 2007.

\bibitem{franklin1996agent}
Stan Franklin and Art Graesser.
\newblock Is it an agent, or just a program?: A taxonomy for autonomous agents.
\newblock In {\em International workshop on agent theories, architectures, and languages}, pages 21--35. Springer, 1996.

\bibitem{maes1993modeling}
Pattie Maes.
\newblock Modeling adaptive autonomous agents.
\newblock {\em Artificial life}, 1(1\_2):135--162, 1993.

\bibitem{wooldridge1995intelligent}
Michael Wooldridge and Nicholas~R Jennings.
\newblock Intelligent agents: Theory and practice.
\newblock {\em The knowledge engineering review}, 10(2):115--152, 1995.

\bibitem{katz1999integrating}
Boris Katz, Deniz Yuret, Jimmy Lin, Sue Felshin, Rebecca Schulman, Adnan Ilik, Ali Ibrahim, and Philip Osafo-Kwaako.
\newblock Integrating web resources and lexicons into a natural language query system.
\newblock In {\em Proceedings IEEE International Conference on Multimedia Computing and Systems}, volume~2, pages 255--261. IEEE, 1999.

\bibitem{wang2024survey}
Lei Wang, Chen Ma, Xueyang Feng, Zeyu Zhang, Hao Yang, Jingsen Zhang, Zhiyuan Chen, Jiakai Tang, Xu~Chen, Yankai Lin, et~al.
\newblock A survey on large language model based autonomous agents.
\newblock {\em Frontiers of Computer Science}, 18(6):186345, 2024.

\bibitem{liu2023chain}
Hao Liu, Carmelo Sferrazza, and Pieter Abbeel.
\newblock Chain of hindsight aligns language models with feedback.
\newblock {\em arXiv preprint arXiv:2302.02676}, 2023.

\bibitem{park2023generative}
Joon~Sung Park, Joseph~C O’Brien, Carrie~J Cai, Meredith~Ringel Morris, Percy Liang, and Michael~S Bernstein.
\newblock Generative agents: Interactive simulacra of human behavior. arxiv.
\newblock {\em arXiv preprint ArXiv:2304.03442}, 2023.

\bibitem{kovavc2023socialai}
Grgur Kova{\v{c}}, R{\'e}my Portelas, Peter~Ford Dominey, and Pierre-Yves Oudeyer.
\newblock The socialai school: Insights from developmental psychology towards artificial socio-cultural agents.
\newblock {\em arXiv preprint arXiv:2307.07871}, 2023.

\bibitem{cui2023chatlaw}
Jiaxi Cui, Zongjian Li, Yang Yan, Bohua Chen, and Li~Yuan.
\newblock Chatlaw: Open-source legal large language model with integrated external knowledge bases.
\newblock {\em CoRR}, 2023.

\bibitem{hamilton2023blind}
Sil Hamilton.
\newblock Blind judgement: Agent-based supreme court modelling with gpt.
\newblock {\em arXiv preprint arXiv:2301.05327}, 2023.

\bibitem{ziems2024can}
Caleb Ziems, William Held, Omar Shaikh, Jiaao Chen, Zhehao Zhang, and Diyi Yang.
\newblock Can large language models transform computational social science?
\newblock {\em Computational Linguistics}, 50(1):237--291, 2024.

\bibitem{bail2024can}
Christopher~A Bail.
\newblock Can generative ai improve social science?
\newblock {\em Proceedings of the National Academy of Sciences}, 121(21):e2314021121, 2024.

\bibitem{canty2024reproducibility}
Richard~B Canty and Milad Abolhasani.
\newblock Reproducibility in automated chemistry laboratories using computer science abstractions.
\newblock {\em Nature Synthesis}, pages 1--13, 2024.

\bibitem{beaulieu2017reproducibility}
Brett~K Beaulieu-Jones and Casey~S Greene.
\newblock Reproducibility of computational workflows is automated using continuous analysis.
\newblock {\em Nature biotechnology}, 35(4):342--346, 2017.

\bibitem{harwell2022sierra}
John Harwell and Maria Gini.
\newblock Sierra: A modular framework for research automation and reproducibility.
\newblock {\em arXiv preprint arXiv:2208.07805}, 2022.

\bibitem{ifargan2025autonomous}
Tal Ifargan, Lukas Hafner, Maor Kern, Ori Alcalay, and Roy Kishony.
\newblock Autonomous llm-driven research—from data to human-verifiable research papers.
\newblock {\em NEJM AI}, 2(1):AIoa2400555, 2025.

\bibitem{besser2018version}
Lilah Besser, Walter Kukull, David~S Knopman, Helena Chui, Douglas Galasko, Sandra Weintraub, Gregory Jicha, Cynthia Carlsson, Jeffrey Burns, Joseph Quinn, et~al.
\newblock Version 3 of the national alzheimer’s coordinating center’s uniform data set.
\newblock {\em Alzheimer Disease \& Associated Disorders}, 32(4):351--358, 2018.

\bibitem{kiselica2020development}
Andrew~M Kiselica, Alyssa~N Kaser, Troy~A Webber, Brent~J Small, and Jared~F Benge.
\newblock Development and preliminary validation of standardized regression-based change scores as measures of transitional cognitive decline.
\newblock {\em Archives of Clinical Neuropsychology}, 35(7):1168--1181, 2020.

\bibitem{kiselica2020using}
Andrew~M Kiselica, Troy~A Webber, and Jared~F Benge.
\newblock Using multivariate base rates of low scores to understand early cognitive declines on the uniform data set 3.0 neuropsychological battery.
\newblock {\em Neuropsychology}, 34(6):629, 2020.

\bibitem{kiselica2023examining}
Andrew~M Kiselica, Ellen Johnson, Kaleea~R Lewis, and Kate Trout.
\newblock Examining racial disparities in the diagnosis of mild cognitive impairment.
\newblock {\em Applied Neuropsychology: Adult}, 30(6):749--756, 2023.

\bibitem{saari2020network}
Toni~T Saari, Ilona Hallikainen, Taina Hintsa, and Anne~M Koivisto.
\newblock Network structures and temporal stability of self-and informant-rated affective symptoms in alzheimer's disease.
\newblock {\em Journal of affective disorders}, 276:1084--1092, 2020.

\bibitem{kurasz2020ethnoracial}
Andrea~M Kurasz, Glenn~E Smith, Maria~G McFarland, and Melissa~J Armstrong.
\newblock Ethnoracial differences in lewy body diseases with cognitive impairment.
\newblock {\em Journal of Alzheimer's Disease}, 77(1):165--174, 2020.

\bibitem{hurst2024gpt}
Aaron Hurst, Adam Lerer, Adam~P Goucher, Adam Perelman, Aditya Ramesh, Aidan Clark, AJ~Ostrow, Akila Welihinda, Alan Hayes, Alec Radford, et~al.
\newblock Gpt-4o system card.
\newblock {\em arXiv preprint arXiv:2410.21276}, 2024.

\bibitem{kapoor2024robust}
Arunima Kapoor, Jean~K Ho, Jung~Yun Jang, and Daniel~A Nation.
\newblock Robust reference group normative data for neuropsychological tests accounting for primary language use in asian american older adults.
\newblock {\em Journal of the International Neuropsychological Society}, 30(4):402--409, 2024.

\end{thebibliography}

\newpage
\section{Appendix}
\label{sec:appendix}

\subsection{Appendix A: Prompt for Study Reproduction}
\label{app:prompt}
\subsubsection{Prompt}
Reproduce the results of the following study: \textit{Examining racial disparities in the diagnosis of mild cognitive impairment}.

\noindent\textbf{Methods:}

\paragraph{Sample}
We requested data through the National Alzheimer’s Coordinating We requested data through the National Alzheimer’s Coordinating Center online portal in November of 2020. Data were provided on November 19, 2020 and contained information dating to the September 2020 data freeze. The initial database included 43,343 participants with baseline data available. Because many of the measures in the UDS include a language component, the database was restricted to English speaking individuals (n ¼ 39,673). The sample was then limited to individuals receiving the most recent version of the UDS neuropsychological battery, version 3.0 (UDS3NB), at baseline (n ¼ 9,564). Next, since we were interested in studying individuals at risk for MCI, the sample was limited to individuals ages 50 and older (n ¼ 9,264). While MCI is typically seen in old age, increasing evidence suggests a high prevalence in middle-age individuals as well (e.g., Kremen et al., 2014), justifying this approach. We then restricted the sample to participants without dementia (i.e., only individuals with CDR global score <1 were included in the sample) because we were interested in deriving normative data in a cognitively unimpaired group and applying this data to better understand diagnosis of MCI (n ¼ 7,711). Finally, because we wanted to make comparisons across non-Hispanic Black/African American and White groups, we restricted the sample to individuals self-identifying as such. The final sample included baseline data for 7,201 participants from 30 Alzheimer’s Disease Research Centers, with study visits occurring from March 2015 through August 2020. The cognitively normal individuals (CDR ¼ 0; n ¼ 3,892) formed a normative sample, from which demographicallyadjusted z-scores were derived (see below for details on how z-scores were calculated). Inferential analyses were performed on the sample of individuals diagnosed with MCI (CDR ¼ .5; n ¼ 3,309). The research was conducted in accordance with the Helsinki Declaration and the guidelines of the University of Missouri Institutional Review Board.

\paragraph{Neuropsychological Measures}
Measures for the UDS 3.0 have been described in detail elsewhere (Besser et al., 2018; Weintraub et al., 2018). For the current study, we used seven core tests, which yield 12 scores (Kiselica et al., 2020). They included (1) a learning/ memory measure, the Craft Story (Craft et al., 1996); (2) two measures of language—a semantic fluency (animal and vegetable trials) task and the Multilingual Naming Test (Gollan et al., 2012; Ivanova et al., 2013), a confrontation naming test; (3) a measure of visual construction and recall, the Benson Figure (Possin et al., 2011); (4) a measure of attention, the Number Span Task (includes forward and backward digit repetition); and (5) a measure of processing speed/executive functioning, the Trail Making Test parts A \& B (Partington \& Leiter, 1949). Actuarial MCI diagnosis Actuarial MCI diagnoses were made according to procedures developed by (Oltra-Cucarella et al., 2018), who defined a base rate of low scores approach to MCI diagnosis. They found this approach to be superior to other methods (e.g., Petersen criteria, Jak/Bondi criteria) in predicting progression to dementia due to Alzheimer’s disease. In the base rate of low scores approach, the number of scores below a certain cutoff is assessed within a normative sample. Then the number of low scores that best approximates the bottom 10\% of the normal curve is set as the cutoff for evidence of objective cognitive impairment.

\paragraph{Actuarial MCI diagnosis}
Actuarial MCI diagnoses were made according to procedures developed by (Oltra-Cucarella et al., 2018), who defined a base rate of low scores approach to MCI diagnosis. They found this approach to be superior to other methods (e.g., Petersen criteria, Jak/Bondi criteria) in predicting progression to dementia due to Alzheimer’s disease. In the base rate of low scores approach, the number of scores below a certain cutoff is assessed within a normative sample. Then the number of low scores that best approximates the bottom 10\% of the normal curve is set as the cutoff for evidence of objective cognitive impairment.

\paragraph{Low Scores}
We defined low scores as those falling below the 9th percentile (z = -1.3408), which is consistent with scores falling below the average range, per the American Academy of Clinical Neuropsychology consensus statement on uniform labeling of scores (Guilmette et al., 2020). We followed procedures from Weintraub et al. (2018) to create demographically adjusted normative scores for each neuropsychological test. Specifically, in the subsample of individuals classified as cognitively normal (CDR global score = 0), each test score was regressed onto age, sex, education, and race. Results of regression analyses are provided in Supplemental Table 1. The subsample included 80\% individuals identifying as White and 20\% of individuals identifying as Black/African American. Regression weights and standard errors of the estimates were then used to create regression-based z-scores (Shirk et al., 2011). Finally, scores were categorized as low if z was less than -1.3408.

\paragraph{Base rates of low scores and actuarial MCI}
Base rates for numbers of low scores for the normative group are presented in Supplemental Table 2. The number of low scores that best approximated the bottom 10\% of the normal distribution was 2+. Thus, individuals were categorized as meeting actuarial criteria for MCI if they had two or more demographically-adjusted z-scores below the 9th percentile.

\paragraph{Amnestic and Non-Amnestic MCI}
Although not a primary focus of the paper, MCI can be broken down into subcategories based on symptom presentation, with the most common distinction being between amnestic and non-amnestic MCI (Petersen, 2004). Amnestic MCI is characterized by cognitive impairment in the memory domain, whereas non-amnestic MCI is characterized by cognitive impairment in domains other than memory (e.g., language or executive functioning). Individuals were classified with amnestic MCI if they had two or more low scores across the neuropsychological battery and at least two of the low scores occurred on memory tests (Craft Story, Benson Figure recall). Individuals were classified as non-amnestic MCI if they had two or more low scores and no more than one low score on a memory test. The amnestic versus non-amnestic distinction was used in post hoc analyses, which are presented in the discussion.

\paragraph{Analyses}
In the subsample of individuals diagnosed as MCI by the CDR (global score = .5), we assessed rates of low scores by race on each test in the UDS3NB using chi-square tests of independence. We also examined racial differences in the overall number of low scores by race using an independent samples t-test. Finally, we tested for racial differences in probability of meeting actuarial criteria for MCI by race using a chi-square test of independence. We expected results to support hypothesis 3: Clinical interview-based methods result in under diagnosis for White persons and/or over diagnosis for Black persons. This hypothesis would be supported if White individuals diagnosed with MCI based on clinical interview demonstrated a higher probability of meeting actuarial MCI criteria than Black individuals.

\subsubsection{Abstract}
Black individuals are less likely to receive an accurate diagnosis of mild cognitive impairment (MCI) than their White counterparts, possibly because diagnoses are typically made by a physician, often without reference to objective neuropsychological test data. We examined racial differences in actuarial MCI diagnoses among individuals diagnosed with MCI via semi-structured clinical interview (the Clinical Dementia Rating) to examine for possible biases in the diagnostic process. Participants were drawn from the National Alzheimer’s Coordinating Center Uniform Data Set and included 491 individuals self-identifying as Black and 2,818 individuals self-identifying as White. Chi-square tests were used to examine racial differences in rates of low scores for each cognitive test (domains assessed included attention, processing speed/executive functioning, memory, language, and visual skills). Next, we tested for racial differences in probability of meeting actuarial criteria for MCI by race. Compared to Black participants diagnosed with MCI via clinical interview, White individuals diagnosed with MCI via clinical interview demonstrated significantly higher rates of low demographically-adjusted z-scores on tests of memory, attention, processing speed, and verbal fluency. Furthermore, White individuals were significantly more likely to meet actuarial criteria for MCI (71.60\%) than Black individuals (57.90\%). Results suggest there may be bias in MCI classification based on semi-structured interview, leading to over diagnosis among Black individuals and/or under diagnosis among White individuals. Examination of neuropsychological test data and use of actuarial approaches may reduce racial disparities in the diagnosis of MCI. Nonetheless, issues related to race-based norming and differential symptom presentations complicate interpretation of results.

\subsubsection{Relevant Columns}

\lstset{
  basicstyle=\ttfamily\small,
  breaklines=true,
  breakatwhitespace=false,
  showstringspaces=false,
  columns=fullflexible,
  xleftmargin=1em
}

\begin{lstlisting}
Here are relevant columns you may use: 
Column: "NACCAGE" - Subject's age at visit. Form: "Subject Demographics" 
Column: "VISITDAY" - Day of visit (1-31). Form: "Form Header and Milestones Form" 
Column: "VISITMO" - Month of visit (1-12). Form: "Form Header and Milestones Form" 
Column: "VISITYR" - Year of visit (eg, 2024). Form: "Form Header and Milestones Form" 
Column: "NACCVNUM" - UDS visit number (order). Form: "Form Header and Milestones Form" 
Column: "NACCETPR" - Primary etiologic diagnosis (MCI; impaired, not MCI; or dementia). Form: "Clinician Diagnosis" 
Possible values: 1 - "Alzheimer-s disease (AD)" 2 - "Lewy body disease (LBD)" 3 - "Multiple system atrophy (MSA)" 
4 - "Progressive supranuclear palsy (PSP)" 5 - "Corticobasal degeneration (CBD)" 
6 - "with motor neuron disease (e.g., ALS)" 7 - ", other" 8 - "Vascular brain injury or vascular dementia including stroke" 
9 - "Essential tremor" 10 - "Down syndrome" 11 - "Huntington-s disease" 12 - "Prion disease (CJD, other)" 
13 - "Traumatic brain injury (TBI)" 14 - "Normal-pressure hydrocephalus (NPH)" 15 - "Epilepsy" 
16 - "CNS neoplasm" 17 - "Human immunodeficiency virus (HIV)" 18 - "Other neurologic, genetic, or infectious condition" 
19 - "Depression" 20 - "Bipolar disorder" 21 - "Schizophrenia or other psychosis" 
22 - "Anxiety disorder" 23 - "Delirium" 24 - "Post-traumatic stress disorder (PTSD)" 
25 - "Other psychiatric disease" 26 - "Cognitive impairment due to alcohol abuse" 
27 - "Cognitive impairment due to other substance abuse" 28 - "Cognitive impairment due to systemic disease or medical illness" 
29 - "Cognitive impairment due to medications" 30 - "Cognitive impairment for other specified reasons (i.e., written-in values)" 
88 - "Not applicable, not cognitively impaired" 99 - "Missing/unknown"

Column: "PRIMLANG" - Primary language. Form: "Subject Demographics" 
Possible values: 1 - "English" 2 - "Spanish" 3 - "Mandarin" 4 - "Cantonese" 
5 - "Russian" 6 - "Japanese" 8 - "Other primary language" 9 - "Unknown"

Column: "NACCUDSD" - Cognitive status at UDS visit. Form: "Clinician Diagnosis" 
Possible values: 1 - "Normal cognition" 2 - "Impaired-not-MCI" 3 - "MCI" 4 - "Dementia"

Column: "CDRGLOB" - Global CDR. Form: "Dementia Staging Instrument" 
Possible values: 0.0 - "No Impairment" 0.5 - "Questionable Impairment" 
1.0 - "Mild Impairment" 2.0 - "Moderate Impairment" 3.0 - "Severe Impairment"

Column: "SEX" - Subject's sex. Form: "Subject Demographics" 
Possible values: 1 - "Male" 2 - "Female"

Column: "EDUC" - Years of education 0-36. Form: "Subject Demographics" 

Column: "NACCNIHR" - Derived NIH race definitions. Form: "Subject Demographics" 
Possible values: 1 - "White" 2 - "Black or African American" 3 - "American Indian or Alaska Native" 
4 - "Native Hawaiian or Pacific Islander" 5 - "Asian" 6 - "Multiracial" 99 - "Unknown or ambiguous"

Column: "FORMVER" - Form version number. Form: "Form Header and Milestones Form" 

Column: "NACCALZD" - NACC Derived variable Presumptive etiologic diagnosis of the cognitive disorder - Alzheimer-s disease. 
Form: "Clinician Diagnosis" 
Possible values: 0 - "No (assumed assessed and found not present)" 1 - "Yes" 8 - "No cognitive impairment"

Column: "TRAILA" - Trail Making Test Part A - Total number of seconds to complete 0 - 150. 
Form: "Neuropsychological Battery Scores" 
Possible values: 995 - "Physical problem" 996 - "Cognitive/behavior problem" 997 - "Other problem" 
998 - "Verbal refusal" -4 - "Not available"

Column: "TRAILB" - Trail Making Test Part B - Total number of seconds to complete. 
Form: "Neuropsychological Battery Scores" 
Possible values: 995 - "Physical problem (Form C1) (Form C1) (Form C1 or C2)" 
996 - "Cognitive/behavior problem" 997 - "Other problem" 998 - "Verbal refusal" -4 - "Not available"

Column: "ANIMALS" - Animals - Total number of animals named in 60 seconds 0-77. 
Form: "Neuropsychological Battery Scores" 
Possible values: 95 - "Physical problem" 96 - "Cognitive/behavior problem" 
97 - "Other problem" 98 - "Verbal refusal" -4 - "Not available"

Column: "VEG" - Vegetables - Total number of vegetables named in 60 seconds 0-77. 
Form: "Neuropsychological Battery Scores" 
Possible values: 95 - "Physical problem" 96 - "Cognitive/behavior problem" 
97 - "Other problem" 98 - "Verbal refusal" -4 - "Not available"

Column: "CRAFTVRS" - Craft Story 21 Recall (Immediate) - Total story units recalled, verbatim scoring 0-44. 
Form: "Neuropsychological Battery Scores" 
Possible values: 95 - "Physical problem" 96 - "Cognitive/behavior problem" 
97 - "Other problem" 98 - "Verbal refusal" -4 - "Not available"

Column: "CRAFTDVR" - Craft Story 21 Recall (Delayed) - Total story units recalled, verbatim scoring 0-44. 
Form: "Neuropsychological Battery Scores" 
Possible values: 95 - "Physical problem" 96 - "Cognitive/behavior problem" 
97 - "Other problem" 98 - "Verbal refusal" -4 - "Not available"

Column: "UDSBENTC" - Total Score for copy of Benson figure 0 - 17. 
Form: "Neuropsychological Battery Scores" 
Possible values: 95 - "Physical problem" 96 - "Cognitive/behavior problem" 
97 - "Other problem" 98 - "Verbal refusal" -4 - "Not available"

\end{lstlisting}

\subsubsection{Ground Rules}
\lstset{
  basicstyle=\ttfamily\small,
  breaklines=true,
  breakatwhitespace=false,
  showstringspaces=false,
  columns=fullflexible,
  xleftmargin=1em
}

\begin{lstlisting}
The dataset can be found at ../data/investigator_nacc56.csv.

The field "NACCID" is a unique patient identifier. The same patients can appear in multiple rows over time.

The field "NACCVNUM" represents the sequential count a visit represents. If methods specify that something occurs in a baseline visit, be sure to filter for 'NACCVNUM == 1'.

If a study specifies that, for example, a neuropsychological battery or other form be completed for partipants, you can check that a given field has data as a proxy for this (eg, 'data['DEL'].isin([0, 1]))').

Be sure to match the study inclusion and exclusion criteria EXACTLY, using the appropriate columns and values, including appropriate handling of not applicable (often '8'), unknown (often '9'), not available (-4) values, or otherwise invalid (often values > 900).

The fields "DRUG1" though "DRUG40" are free-text fields which contain names of drugs patients are taking. Thus to find patients taking a drug such as metformin, you would need to search all "DRUG1", "DRUG2", and so on fields for relevant drugs as strings. Also, drug names in the dataset are all upper-cased (eg 'METFORMIN'), so be sure to take this into consideration for analysis.

If a study says participants taking certain categories of drugs were included/excluded, but does not name the specific drugs, reason about what those drugs fit those categories and search for names of drugs you know.

You need ONLY REPRODUCE THE RESULTS DESCRIBED IN THE ABSTRACT. Do not attempt to reproduce other aspects of the study described in Methods.

You can compute a 'VISIT_DATE' column by something like 'data['VISIT_DATE'] = pd.to_datetime(data['VISITDAY'].astype(str) + '-' + data['VISITMO'].astype(str) + '-' + data['VISITYR'].astype(str), errors='coerce')'

Be sure to IGNORE VALUES WHICH COULD RESULT IN ERRONEOUS ANALYSIS, such as not applicable (often '8'), unknown (often '9'), and not available ('-4'). Double-check erroneous value types (which may differ column-to-column) are filtered out appropriately.

If you do not know the columns needed to reproduce a portion of the study, ignore that aspect.

Even if a given analysis was conducted in R, write and execute your code in equivalent Python.

Be sure to output patient counts and use well-documented print() statements at each step. If study Methods provide patients counts for each inclusion/exclusion criteria (eg 'n = 1,000'), be sure your output follows this pattern for analysis purposes (and ensuring you count only unique NACCIDs, ie individual patients and not visits, unless otherwise specified).

The data you have available may be slightly newer than those used in the study, and thus patient counts may be slightly higher at each step than those published.

Be sure the inclusion and exclusion criteria STRICTLY follow the steps laid out in the study Methods.

Do not attempt to produce visualizations.

Your goal is to determine whether you can reproduce the Results of the study outlined in the Abstract.
\end{lstlisting}

\subsubsection{Example Data}
Here is as example of NACC data:
\begin{verbatim}
NACCAGE VISITDAY VISITMO VISITYR NACCVNUM NACCETPR PRIMLANG
68      14       9        2021     1        88         1
70      16       11       2021     1        88         1
72      21       9        2021     1        88         1

NACCUDSD CDRGLOB SEX EDUC NACCNIHR FORMVER NACCALZD TRAILA TRAILB
1         0.5     2   14    1        3.0     8         32     68
1         0.0     2   14    2        3.0     8         34     84
1         0.0     1   14    1        3.0     8         29     76

ANIMALS VEG CRAFTVRS CRAFTDVR UDSBENTC
22      11  31        22        16
16      14  16         9        15
19      11  29        29        16
\end{verbatim}

\end{document}